\documentclass[10pt,twocolumn,letterpaper]{article}

\usepackage{iccv}
\usepackage{times}
\usepackage{epsfig}
\usepackage{graphicx}
\usepackage{amsmath}
\usepackage{amssymb}
\usepackage{booktabs}
\usepackage{cuted}
\usepackage{makecell}
\usepackage{caption}

\usepackage[pagebackref=true,breaklinks=true,letterpaper=true,colorlinks,bookmarks=false]{hyperref}

\iccvfinalcopy 

\def\iccvPaperID{3635} 

\ificcvfinal\fi

\def\methodName{SceneGenie}
\newcommand{\xpm}[1]{{\tiny$\pm#1$}}

\begin{document}

\title{\methodName{}: Scene Graph Guided Diffusion Models for Image Synthesis}

\author{Azade Farshad${}^{1,2}$ \thanks{The first two authors contributed equally to this work} \hspace{0.9cm} 
Yousef Yeganeh${}^{1,2}$ \footnotemark[1]  \hspace{0.9cm}
Yu Chi${}^{1}$  \hspace{0.9cm} 
Chengzhi Shen${}^{1}$   \\ \vspace{-0.25cm} \\ 
Bj{\"o}rn Ommer${}^{3}$ \hspace{0.9cm} 
Nassir Navab${}^{1}$ 
\and
${}^{1}$Technical University of Munich
\and
${}^{2}$MCML
\and
${}^{3}$Ludwig Maximilian University
}

\maketitle
\ificcvfinal\thispagestyle{empty}\fi

\begin{strip}
\begin{center}
\centering
\vspace{-2cm} 
\resizebox{\linewidth}{!}{
 \begin{tabular}{c@{\hskip 0.15cm}c@{\hskip 0.15cm}c@{\hskip 0.15cm}c@{\hskip 0.15cm}}
\multicolumn{4}{c}{\textbf{Prompt}: a \underline{sheep} by another \underline{sheep} standing on the \underline{grass} with \underline{sky} above and the \underline{ocean} by a \underline{tree} and a \underline{boat} on the \underline{grass}} \vspace{0.15cm}\\

\raisebox{-.5\height}{\includegraphics[width=0.25\textwidth]{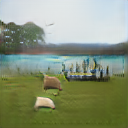}} &
 \raisebox{-.5\height}{\includegraphics[width=0.25\textwidth]{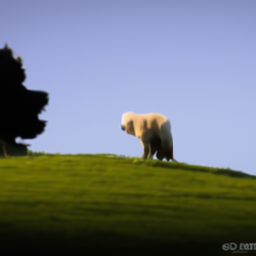}} &
    \raisebox{-.5\height}{\includegraphics[width=0.25\textwidth]{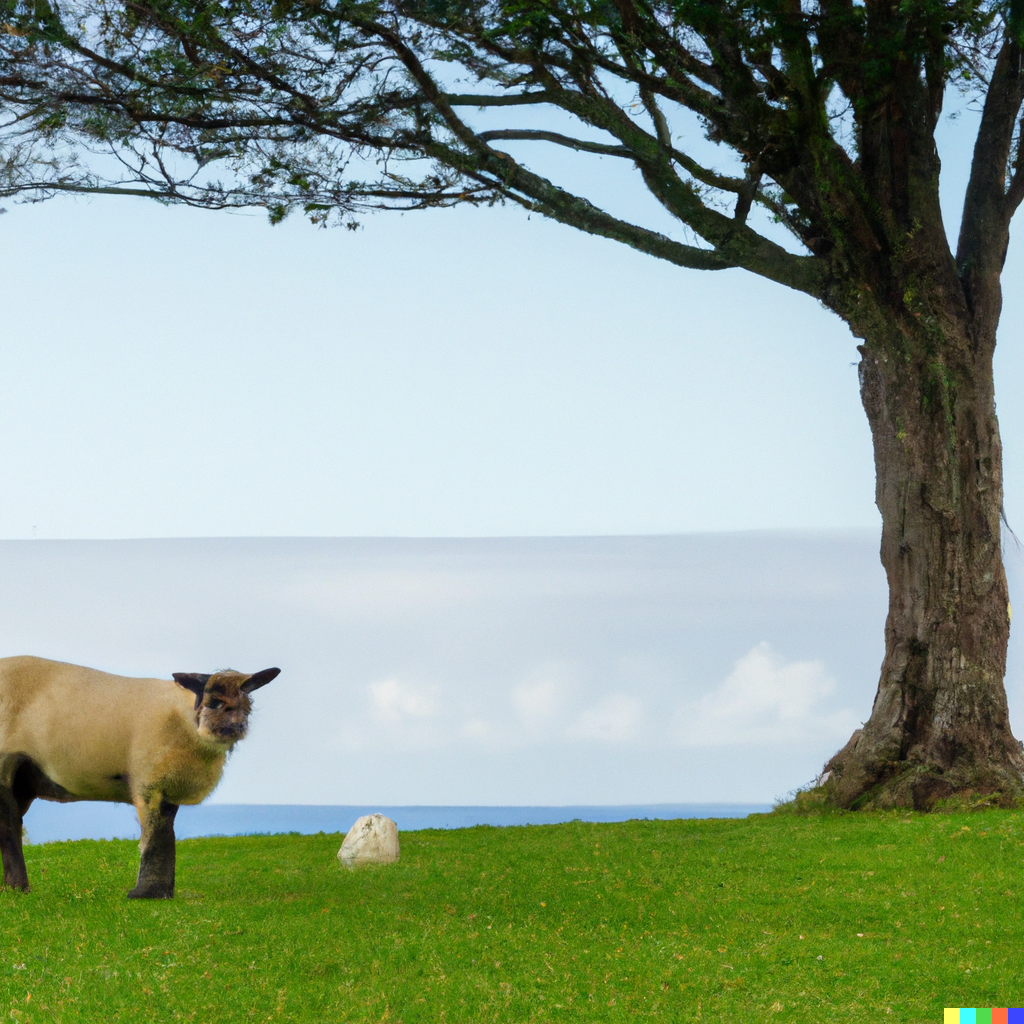}} & 
 \raisebox{-.5\height}{\includegraphics[width=0.25\textwidth]{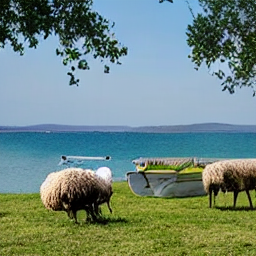}}
    \vspace{0.15cm} \\
SG2Im \cite{johnson2018image} & GLIDE \cite{nichol2021glide} & DALL.E 2 \cite{ramesh2022hierarchical} & \methodName{} (Ours)\vspace{0.15cm} \\
 \end{tabular}
}
\captionof{figure}{Synthesized images from SG2Im \cite{johnson2018image}, GLIDE \cite{nichol2021glide}, DALL.E 2 \cite{ramesh2022hierarchical}, and \methodName{}. For SG2Im, and \methodName{}, the sentence is first converted to a scene graph before feeding into the model. While the sentence describes two sheep and a boat in a specified scene, text-to-image generators like GLIDE, and even larger models like DALL.E 2 generate considerably inaccurate results; while \methodName{} accurately represents the scene defined by the prompt.} 
\label{ch:res:visualization}

\end{center}
\end{strip}

\begin{abstract}
Text-conditioned image generation has made significant progress in recent years with generative adversarial networks and more recently, diffusion models. While diffusion models conditioned on text prompts have produced impressive and high-quality images, accurately representing complex text prompts such as the number of instances of a specific object remains challenging.

To address this limitation, we propose a novel guidance approach for the sampling process in the diffusion model that leverages bounding box and segmentation map information at inference time without additional training data.  Through a novel loss in the sampling process, our approach guides the model with semantic features from CLIP embeddings and enforces geometric constraints, leading to high-resolution images that accurately represent the scene. To obtain bounding box and segmentation map information, we structure the text prompt as a scene graph and enrich the nodes with CLIP embeddings. Our proposed model achieves state-of-the-art performance on two public benchmarks for image generation from scene graphs, surpassing both scene graph to image and text-based diffusion models in various metrics. Our results demonstrate the effectiveness of incorporating bounding box and segmentation map guidance in the diffusion model sampling process for more accurate text-to-image generation.
\end{abstract}

\section{Introduction}
\label{sec:intro}
Image generation using deep neural networks is a rapidly evolving field in computer vision, with the objective of creating models that have a deep understanding of the objects and scenes they are creating. In recent years, significant progress has been made in text-to-image synthesis using Recurrent Neural Networks (RNNs) \cite{zia2020text} and Generative Adversarial Networks (GANs) \cite{park2019semantic, sun2019image}, which can generate high-quality, photorealistic images from textual descriptions. Lately, diffusion models, a class of generative models, excelled GAN models \cite{dhariwal2021diffusion} and became the prominent method in the image generation task. However, most of these methods often struggle with creating complex scenes from long, natural language descriptions. This is because sentences are linear structures that may not efficiently describe complex scenes. 

To tackle this problem, we propose \methodName{}, which is a novel layout-based approach for guiding the sampling process of a diffusion model. Our method leverages bounding box and segmentation information as a guidance in the reverse sampling process. The bounding box and segmentation map information are predicted by a Graph Neural Network (GNN) after structuring the text prompt in the form of a scene graph. We propose using scene graphs as they are powerful structured representations of objects and their relationships in both the image and language domains.

Our proposed guidance is similar to classifier guidance. We compute the classifier gradients for each object based on the distance between the CLIP image embedding in the region of interest (RoI) for that specific object and the corresponding CLIP text embedding for that object in the form of \textit{a Photo of an \textbf{obj}}. To compute object-wise gradients in the RoI, we inject gaussian noise outside the RoI and then compute the total gradient as the weighted sum of the gradients for different objects in the scene. For segmentation guidance, we take advantage of the first-stage autoencoder of the diffusion model to measure how semantically close the segmentation map and the image are. Using such guidance for the diffusion model sampling results in higher quality and more accurate images that better represent the input prompt.

Recently, there have been approaches such as Make a Scene \cite{gafni2022make,xu2022hierarchical} that condition the diffusion model directly on the segmentation map or scene layout, or methods such as SDEdit \cite{meng2021sdedit} that use the segmentation map as the initialization in the sampling process. However, these methods either require additional training for the input condition or need different architectural designs for different conditions.

Our proposed method differs from these other works in that it directly optimizes the sampling process using additional information and does not necessarily need to be paired with the images in the training dataset due to its usage at inference time. This allows us to create more complex and accurate scenes while still maintaining the high quality of generated images. We demonstrate the effectiveness of our approach through experiments on public benchmarks, showing that our method outperforms existing text-to-image diffusion models as well as state-of-the-art scene graph to image approaches without any additional training.
 
In summary, our work makes several key contributions:
\begin{itemize}
    \item We propose a novel approach for guiding the sampling process in a diffusion model that places greater emphasis on the regions of interest (RoI) by incorporating the gradients computed from predicted bounding box and segmentation maps. 
    \item Our proposed guidance is applied during the reverse sampling process. Therefore, it does not require any additional training and can be applied to any diffusion model architecture.
    \item To enable the use of bounding box guidance in the sampling process, we propose a novel method of noise injection outside the RoI. For the segmentation guidance, we take advantage of the first-stage autoencoder of the diffusion model. Therefore, we effectively leverage the bounding box and segmentation map information and improve the accuracy of generated images.
    \item Our method achieves notably higher image generation performance compared to scene graph to image models in high resolution image generation, and outperforms the state-of-the-art in text-to-image generation.
    \item Finally, we demonstrate that incorporating CLIP embeddings as node features in the scene graph improves the accuracy of bounding box and segmentation predictions.
    
\end{itemize}

\section{Related Work} \label{sec:rel_work}
The high dimensionality of images poses a challenge for image generation based on deep learning. Recent advances in generative models, in particular, Generative Adversarial Networks (GANs)~\cite{goodfellow2014generative}, have boosted the quality and diversity of generated images. A line of works explore generative models for unconditional image generation~\cite{karras2020analyzing,karras2021alias}. Conditional image generation models have also been explored \cite{mirza2014conditional} with a diverse set of priors such as semantic segmentation maps~\cite{chen2017photographic,wang2018high,park2019semantic}, natural language descriptions~\cite{zhang2018photographic,li2019storygan} or translating from one image domain to another using paired~\cite{isola2017image} or unpaired data~\cite{zhu2017unpaired}. Conditional image generation models also enable the possibility of interactive image manipulation by partial image generation using hand-crafted part replacement \cite{hu2013patchnet} or by incorporating a user interface for specifying the locations that need to be inpainted \cite{li2022cross}. The inpainting process of the specified regions \cite{pathak2016context,ling2021editgan} can also be guided by semantic information \cite{yeh2017semantic,hong2018learning,ntavelis2020sesame} or edges \cite{Yu_2019_ICCV,nazeri2019edgeconnect}. For instance, in GLIDE \cite{nichol2021glide}, the model is capable of replacing original content with the guidance from the CLIP \cite{radford2021learning} embeddings by extra fine-tuning.

\begin{figure*}
    \centering
    \includegraphics[width=\linewidth]{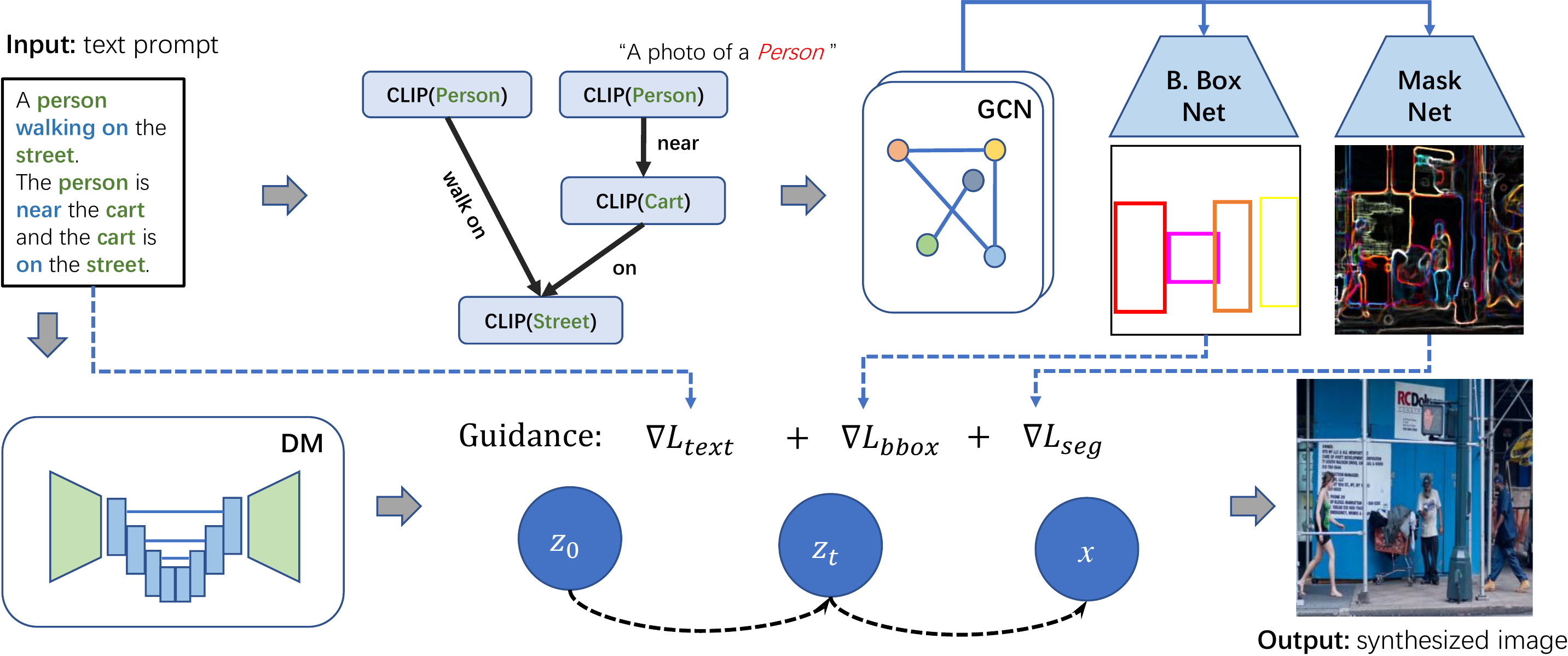}
    \caption{\textbf{Overview of \methodName{}.} Our pipeline starts by getting a text prompt as input. The triplets are structured in form of a scene graph. The graph is then processed by a GCN which outputs object embeddings per node in the graph. The embeddings are used to predict bounding box coordinates and pseudo-segmentation maps corresponding to each object in the scene, as well as a final segmentation map from the whole image. We use a diffusion model to generate images conditioned on the text prompt by guiding the sampling process through the bounding box and segmentation map information.}
    \label{fig:method}
\end{figure*}
\paragraph{Diffusion Models}
A recent and impressive improvement in image generation is achieved by diffusion models \cite{ho2020denoising, dhariwal2021diffusion}. Diffusion models \cite{ho2020denoising, dhariwal2021diffusion} are generative models that produce images by successively denoising images. Unconditional image generation with diffusion models initiated with the denoising approach with the work of Sohl \etal \cite{sohl2015deep} that formalized diffusion models as multi-scale convolutions, and then the same work was extended by Ho \etal \cite{ho2020denoising}; later, Song \etal \cite{song2020denoising} proposed a non-Markovian method for the forward process. Conditional image generation by diffusion models based on classifier guidance was proposed by Dhariwal \etal \cite{dhariwal2021diffusion}, which uses a classifier to guide the diffusion model during the sampling process. In many applications, the conditional diffusion models are utilized; for instance, in Palette \cite{saharia2022palette}, the diffusion model is conditioned on a low-resolution image to generate a high-resolution image, or in SDEdit \cite{meng2021sdedit}, the diffusion model is conditioned with a low-quality image to sharpen and enhance colors and textures. Recently, image generation conditioned on text has captured a lot of attention, where CLIP \cite{radford2021learning} guidance is often utilized.

\paragraph{Contrastive Language-Image Pre-Training}
By successfully incorporating text-and-image pairs through contrastive learning, Contrastive Language-Image Pre-Training (CLIP) \cite{radford2021learning} has been extensively applied in object detection \cite{teng2021global, han2016seq}, image captioning \cite{mokady2021clipcap, hessel2021clipscore}, and text-to-image generation \cite{nichol2021glide}. CLIP \cite{radford2021learning} is able to capture similar representation beyond modality by minimizing the distance between text and picture embeddings from the same pair while maximizing the distance between those from dissimilar pairs. 

\paragraph{Scene Graph to Image}
Scene graphs~\cite{johnson15} are graphs that represent a scene by defining the objects in the scene as nodes in the graph and the relationships between them as edges. Scene graphs gained more attention recently due to the rise of large-scale scene graph datasets such as Visual Genome \cite{krishna2017visual}, MOMA \cite{luo2021moma}, and Action Genome \cite{ji2020action}. A broad line of works ~\cite{li2017scene,xu2017scenegraph,newell2017pixels,herzig2018mapping, qi2018attentive, zellers2018neural,tang2020unbiased,suhail2021energy} explore the generation of scene graphs from images, while Johnson \etal proposed SG2Im \cite{johnson2018image}, which is the first pipeline that attempts to generate images from scene graphs. Typically, the task can be divided into two parts: first, convert a scene graph to an intermediate layout, then use the layout as an input to conditional GANs \cite{chen2017photographic,park2019semantic,sun2019image} for image synthesis. Some works~\cite{zhao2018image,sun2019image,sylvain2021object} focus on generating images directly from scene layouts.
Recent works either focus on enhancing the model's capacity for graph understanding \cite{garg2021unconditional} or improving intermediate layout quality \cite{yang2021layouttransformer}. Herzig \etal \cite{herzig2020learning} deals with semantic equivalence in large complex scene graphs, while \cite{ivgi2021scene} intends to reduce blurry and overlapping objects in the scene layout in a coarse-to-fine manner. Modified GCNs are engaged in a cascaded refinement network in \cite{chen2017photographic} to reduce high-dimensional embeddings. However, the literature has widely studied that deep Graph Neural Networks suffer from over-smoothing issues \cite{oono2019graph, li2018deeper, alon2020bottleneck} that average all information in the graph, causing semantic ambiguities. Recently, there has been models \cite{dhamo2020semantic, mittal2019interactive, ashual2019specifying} proposed that focus on image manipulation, where users are able to control the synthesis results by modifying the scene graph interactively. 
\section{Background}
Diffusion models use a Markov chain that gradually adds Gaussian noise to an image $x_{0}$ to get the approximate posterior $q(x_{1:T}|x_{0})$ with $x_{1},\ldots,x_{T}$ being the noisy versions of $x_{0}$. If $T$ is large enough $x_{T}$ is approximated by $\mathcal{N}(0, \mathcal{I})$. By learning the reverse process $p_{\theta}(x_{t-1}|x_{t}) := \mathcal{N}(\mu_{\theta}(x_{t}), \sum_{\theta}(x_{t}))$ of this Markov chain, one can generate new images $x_{0} \sim p_{\theta}(x_{0})$ from pure noise $x_{T} \sim \mathcal{N}(0, \mathcal{I})$ by gradually denoising in a sequence of steps $x_{T-1}, x_{T-2}, \ldots, x_{0}$.
Such a model is obtained by generating noisy samples $x_{t} \sim q(x_{t}|x_{0})$ and training a model $\theta$ (typically a U-Net) to predict the added noise using an MSE loss:
\begin{equation}
    L_{DM} = E_{x_{0}, \epsilon, t}[\left \| \epsilon - \epsilon_{\theta}(x_{t}) \right\|^{2}]
\end{equation}
This model can then successively generate images by denoising images step by step starting from pure noise. 

\subsection{Guided Diffusion}

Denoising diffusion probabilistic models (DDPM) \cite{ho2020denoising} have shown exceptional performance in unconditional image generation. Yet, generating images with desired semantics is still challenging due to DDPM's nature as a stochastic generation process.
Therefore previous work has focused on classifier guiding \cite{dhariwal2021diffusion}, perturbing the mean $\mu_{\theta}(x_{t}|y)$ and variance $\sum_{\mu}(x_{t}|y)$ of the diffusion model by a classifier gradient. The perturbed mean $\hat{\mu_{\theta}}(x_{t}|y)$ is given by
\begin{equation}\label{eq:1}
\hat{\mu_{\theta}}(x_{t}|y) = \mu_{\theta}(x_{t}|y) + \alpha \cdot\Sigma_{\theta} (x_{t}|y) \nabla x_{t} \log{ p_{\phi}(y|x_{t})} 
\end{equation} Where $\alpha$ is a hyperparameter called guidance scale that controls sample quality vs. sample diversity \cite{dhariwal2021diffusion}.

\subsection{Latent Diffusion Models}
Latent Diffusion Models (LDMs) \cite{rombach2022high} are trained to apply the diffusion process and reverse sampling process on image latent space, which significantly reduces the computational complexity compared to the diffusion models trained on the image space. Latent embeddings of the images are coded by Kl-autoencoder \cite{rombach2022high} or VQGAN \cite{esser2020taming}. The diffusion and denoising processes of LDM \cite{rombach2022high} can be derived as:
\begin{equation}
    q(z_{1:T}|z_{0}) = \Pi_{t=1}^{T}q(z_{t}|z_{t-1})
\end{equation}
\begin{equation}
        p_{\theta}(z_{0:T}) = p(z_{T})\Pi_{t=1}^{T} p_{\theta}(z_{t-1}|z_{t})
\end{equation}

\section{Methodology} \label{sec:method}
Our method consists of two steps: 1) Training a model for the prediction of bounding boxes and segmentation maps from a scene graph obtained from the text prompt, 2) Generating the image guided by the bounding box coordinates, the segmentation map, and the text embedding. We focus on the extraction of bounding boxes and segmentation maps from scene graphs in \autoref{sg2seg}, while \autoref{diffusion} concentrates on guiding the diffusion model in order to generate more accurate images.

We are given a dataset $\mathcal{D}$ of images $x$, and text prompts $\tau$, and bounding box coordinates $c$. Optionally, we can have segmentation maps $s$. The text prompts are split into triplets in the form of (object, predicate, subject), where the predicate $r$ defines the relationship between the object and the subject. The scene graph $\mathcal{G}$ is composed of the the object categories $o$ and the relationships $r$, where object categories are the nodes in the graph and the edges are the relationships between them. More formally, a graph can be formed as $\mathcal{G}=(O, E)$ where 
$O= \left\{ o_1, …, o_n \right\}$ 
are $n$ objects in the graph, and 
$E= \left\{ (o_i, r, o_j) | o_i, o_j \in O, r \in R \right\}$ with $R$ as the relationship category between objects.


\subsection{Scene Graph to Segmentation (SG2SEG)}
\label{sg2seg}
Given the scene graph $\mathcal{G}$ with objects (nodes) and relationships (edges), we aim to synthesize segmentation maps that transform the information from the text space to the image space. These segmentations ought to be realistic in terms of object shapes, as well as semantically consistent in terms of object relationships. 

Firstly, we acquire object embeddings using CLIP \cite{radford2021learning} features in each node of the graph. The assumption is that CLIP \cite{radford2021learning} is able to generate the object features $f_{obj}$ that are consistent with the text feature $f_{text}$ describing each object. To form the input to CLIP's \cite{radford2021learning} text encoder,  we build a prompt for each $o_i \in O$, e.g., \textit{a photo of an [obj]} with \textit{[obj]} substituted by the corresponding object classes. The generated CLIP \cite{radford2021learning} features $f_{text} \in R^{n \times 512}$ are then fed to a Graph Neural Network to learn the object embedding $f_{emb} \in R^{n \times d}$, where $d$ is a hyperparameter that controls the embedding dimensionality. As a trade-off between computation and model expressiveness, we select $d=128$. Note that, we treat edge $E$, that stands for relationships, as learnable embeddings.

Secondly, we explicitly constrain the output of our SG2SEG on ground truth, instead of learning an intermediate representation, as in SG2Im \cite{johnson2018image}. Given the i-th object embedding $f_{emb}^{i} \in R ^{1 \times d}$, we apply a mask regression network (Mask Net) and box regression network (B. Box Net) as in SG2Im \cite{johnson2018image}, but remove the remaining parts of layout sampling and merging. The benefits are twofold: 1) By explicitly constraining the predicted mask and bounding boxes, we are able to achieve higher quality in both results, 2) Predictions are more reliable with even less computational power.

The SG2SEG network is optimized with three objective functions: \\
1) Box loss $L_{box} = \sum_{i=1}^m||c_i, \hat c_i||_1$ is the L1 difference between the 4 coordinate values of the predicted and ground truth bounding boxes. \\
    2) Mask loss $L_{mask} = BCE(m_i, \hat m_i)$ is the binary cross entropy loss for each predicted object mask and ground truth. \\
    3) Segmentation loss $L_{seg} = ||s, \hat s||_1$ is the L1 difference between the segmentation map from predicted masks and the ground truth. It helps the network to generate more consistent results when merging multiple object masks together.

\begin{figure}[ht]
    \begin{center}
      \includegraphics[width=\linewidth]{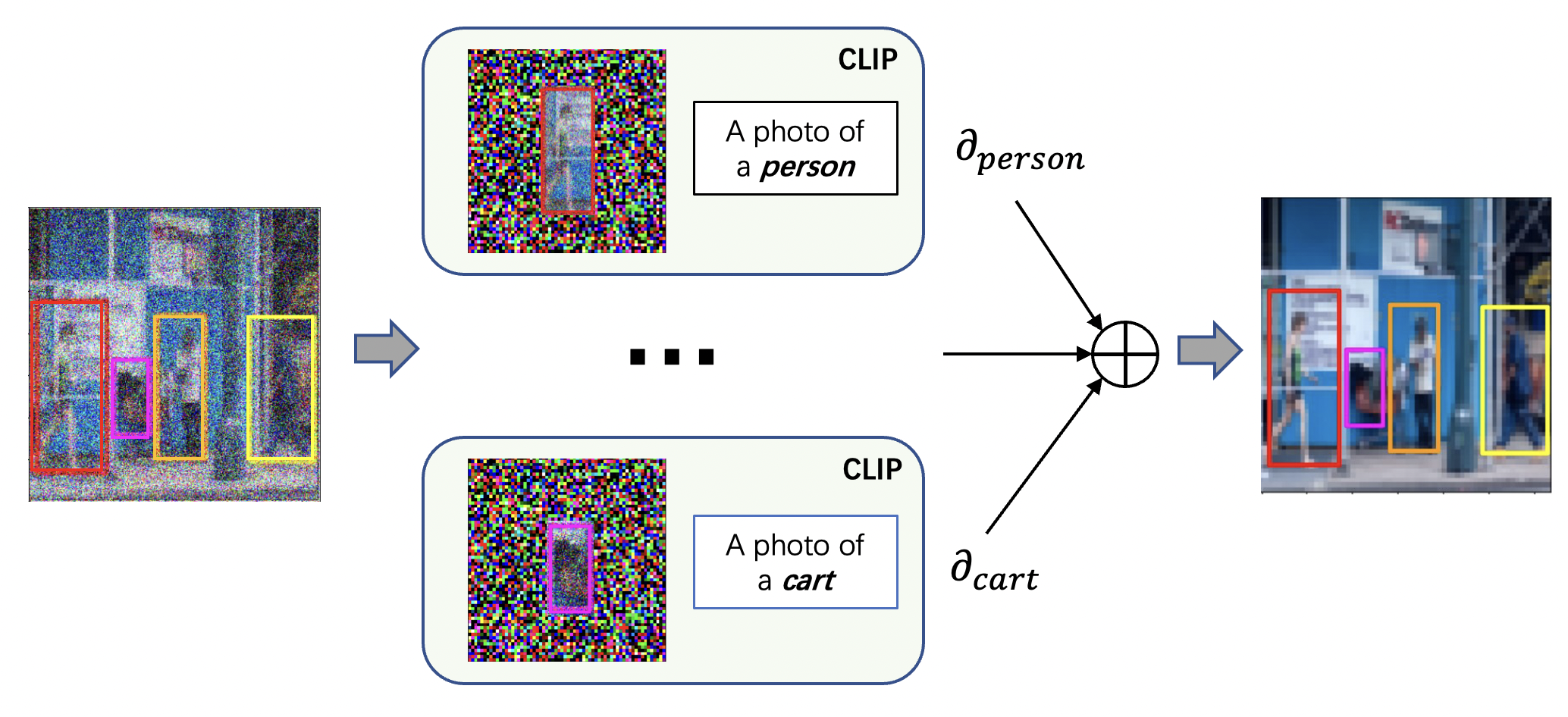} 
    \end{center}
       \caption{Example of Gaussian Noise padded bounding box in the image. We compute the gradients ($\partial obj$) for each object in the image based on the CLIP score and then aggregate the gradients for all the object for the backpropagation step.}
\label{fig:noise_patch}
\end{figure}
\subsection{Conditional Image Generation with Diffusion Models}
\label{diffusion}
In our work, we replace the gradient of the classifier in \autoref{eq:1} with combinations of gradients defined in \autoref{clipg}, \autoref{clipbb}, and \autoref{diffinit}. The main objective is to guide the image generation toward a correct scene layout and object realism. 

\subsubsection{CLIP Text Guidance} \label{clipg}
In order to generate an image which corresponds to a specific input text in the sampling process, we add guidance in the sampling process. Given an input text $text$ and input image $img$ along with CLIP \cite{herzig2020learning} image encoder $E_{i}$ and CLIP \cite{herzig2020learning} text encoder $E_{t}$, we calculate the gradient of the CLIP score with respect to the latent space $z$ of input image of LDM \cite{rombach2022high}. The calculated gradient is used in the sampling process for the guidance as described in equation \ref{eq:1}. It is formulated as:
\begin{equation}
    L_{text} = E_{i}(img)*E_{t}(text)
\end{equation}
For LDM, the gradient is formulated as:
\begin{equation}
    \nabla L_{text} = -\frac{\partial L_{text}}{\partial z}
\end{equation}
For other diffusion model whose diffusion process is on image space $x$, the gradient is computed by:
\begin{equation}
    \nabla L_{text} = -\frac{\partial L_{text}}{\partial x}
\end{equation}

\subsubsection{CLIP Bounding Box Guidance} \label{clipbb}
In addition to using an input prompt to generate the entire image, we want to make a certain region in image that corresponds to the input prompt. Here we propose CLIP \cite{herzig2020learning} Bounding Box Guidance. From SG2SEG framework, the size and location of bounding box of a certain object is inferred.

For an object $obj_{k}$ in a bounding box, we pad the bounding box with Gaussian Noise into the size of the original image as illustrated in \autoref{fig:noise_patch}. Then we calculate the CLIP score between CLIP \cite{radford2021learning} image embedding of Gaussian Noise padded Bounding Box $obj_{k}$ and CLIP text embedding of the category of object $l_{k}$ in format starting with "A photo of" as follows:
\begin{equation}  
 L_{obj_{k}} = E_{i}(obj_{k})*E_{t}(l_{k})
\end{equation} 
where $E_{i}$ and $E_{t}$ are pretrained CLIP Image Encoder and Text Encoder.\\
Assuming there are N objects $obj_{1}, obj_{2},..., obj_{N}$ in one image, we have the entire bounding box guidance score as weighted sum of the object bounding box guidance: 
\begin{equation}
    L_{boxg} = \sum_{i=1}^{N} w_{i}*L_{obj_{i}}
\end{equation}
The weights $w_{i}$ are normalized and proportional to the size of each bounding box.
The gradient used for the guidance of LDM\cite{rombach2022high} will be:
\begin{equation}
    \nabla L_{boxg} = -\frac{\partial L_{boxg}}{\partial z}
\end{equation}
, where $z$ is the image latent of LDM\cite{rombach2022high}.
For other diffusion model whose diffusion process is on image space $x$, the gradient is formulated as 
\begin{equation}
    \nabla L_{boxg} = -\frac{\partial L_{boxg}}{\partial x}
\end{equation}

\begin{figure*} 
\centering
\resizebox{\linewidth}{!}{
 \begin{tabular}{c@{\hskip 0.15cm}c@{\hskip 0.15cm}c@{\hskip 0.15cm}c@{\hskip 0.15cm}c@{\hskip 0.15cm}c@{\hskip 0.15cm}c@{\hskip 0.15cm}}
\textbf{Prompt} &
\thead{A bicycle \\ leaning against \\ a bed in a bedroom} &
\thead{A skier is standing \\ at the top of \\ ski slope} &
\thead{The three teddy bears \\ sit with their arms \\ around each other} &
\thead{Three wild turkeys \\ on top of \\ the dried pasture} &
\thead{A portrait picture \\of a young man \\in a suit and tie} &
\thead{Two people \\standing on a sandy beach \\flying a kite} 
\vspace{0.15cm}\\
 
\rotatebox[origin=c]{90}{\textbf{GT}} & 
\raisebox{-.5\height}{\includegraphics[width=3cm]{figures/examples/1_gt.jpeg}} &
 \raisebox{-.5\height}{\includegraphics[width=3cm]{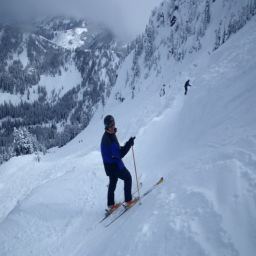}} &
    \raisebox{-.5\height}{\includegraphics[width=3cm]{figures/COCO/gt/296_test.png}}  &
 \raisebox{-.5\height}{\includegraphics[width=3cm]{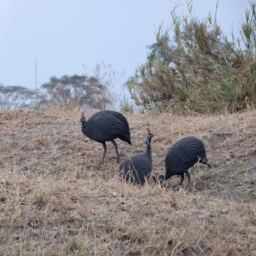}} &
 \raisebox{-.5\height}{\includegraphics[width=3cm]{figures/COCO/gt/408_test.png}} &
\raisebox{-.5\height}{\includegraphics[width=3cm]{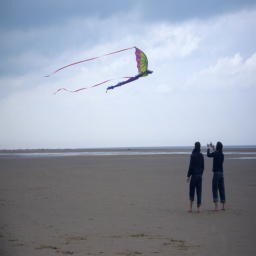}}
    \vspace{0.15cm} \\
 \rotatebox[origin=c]{90}{\textbf{SG2Im \cite{johnson2018image}}}  &
    \raisebox{-.5\height}{\includegraphics[width=3cm]{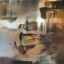}} &
    \raisebox{-.5\height}{\includegraphics[width=3cm]{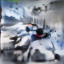}} &
    \raisebox{-.5\height}{\includegraphics[width=3cm]{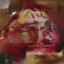}} &
    \raisebox{-.5\height}{\includegraphics[width=3cm]{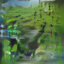}} &
    \raisebox{-.5\height}{\includegraphics[width=3cm]{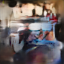}} &
     \raisebox{-.5\height}{\includegraphics[width=3cm]{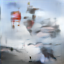}}
    \vspace{0.15cm} \\
 \rotatebox[origin=c]{90}{\textbf{GLIDE \cite{nichol2021glide}}}  &
    \raisebox{-.5\height}{\includegraphics[width=3cm]{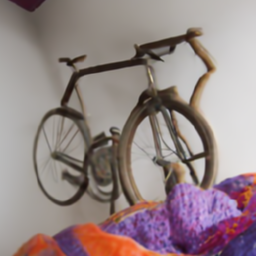}} &
    \raisebox{-.5\height}{\includegraphics[width=3cm]{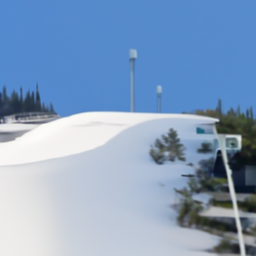}} &
    \raisebox{-.5\height}{\includegraphics[width=3cm]{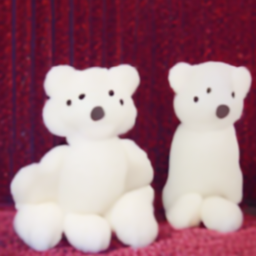}} &
    \raisebox{-.5\height}{\includegraphics[width=3cm]{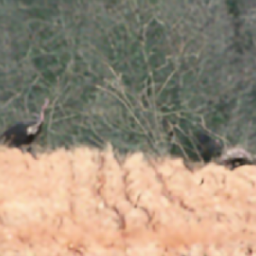}} &
    \raisebox{-.5\height}{\includegraphics[width=3cm]{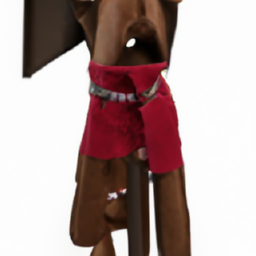}} &
    \raisebox{-.5\height}{\includegraphics[width=3cm]{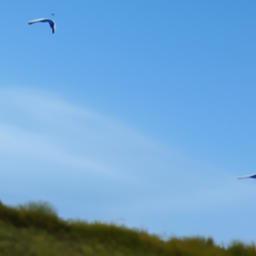}}
    \vspace{0.15cm} \\
\rotatebox[origin=c]{90}{\textbf{LDM \cite{rombach2022high}}}  &
 \raisebox{-.5\height}{\includegraphics[width=3cm]{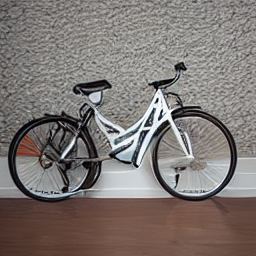}} &
 \raisebox{-.5\height}{\includegraphics[width=3cm]{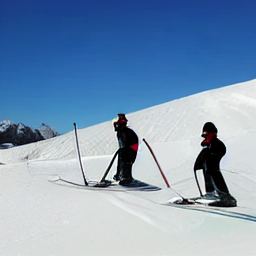}} &
 \raisebox{-.5\height}{\includegraphics[width=3cm]{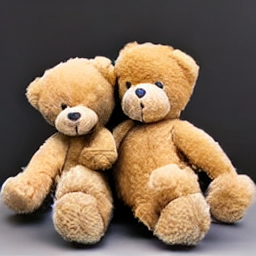}} &
  \raisebox{-.5\height}{\includegraphics[width=3cm]{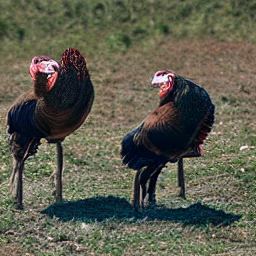}} &
  \raisebox{-.5\height}{\includegraphics[width=3cm]{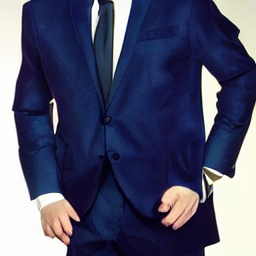}} &
  \raisebox{-.5\height}{\includegraphics[width=3cm]{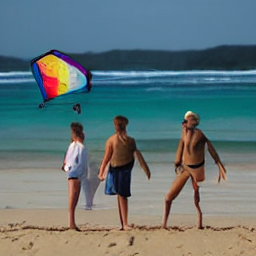}}
 \vspace{0.15cm} \\  
 \rotatebox[origin=c]{90}{\textbf{\methodName{} (Ours)}}  &
    \raisebox{-.5\height}{\includegraphics[width=3cm]{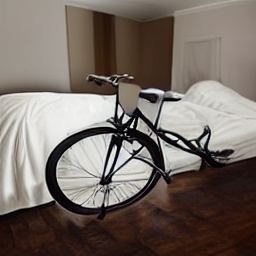}} &
    \raisebox{-.5\height}{\includegraphics[width=3cm]{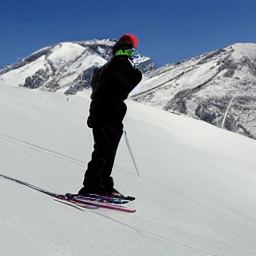}} &
    \raisebox{-.5\height}{\includegraphics[width=3cm]{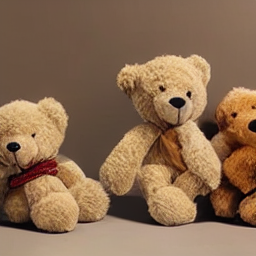}} &
    \raisebox{-.5\height}{\includegraphics[width=3cm]{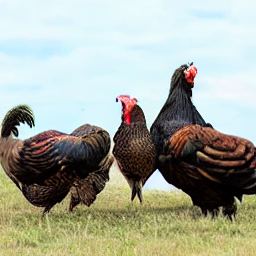}} &
    \raisebox{-.5\height}{\includegraphics[width=3cm]{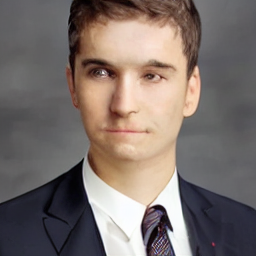}} &
     \raisebox{-.5\height}{\includegraphics[width=3cm]{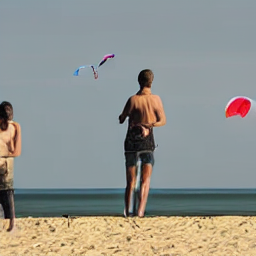}} 
    \vspace{0.15cm} \\
 \end{tabular}
}
    \caption{\textbf{Some qualitative results on the comparison of \methodName{} against related work on the COCO stuff \cite{caesar2018coco} test set.} As it can be seen, the images generated by \methodName{} represent the given prompt more accurately compared to previous work. \methodName{}, in addition to high quality image generation, correctly generates the number of given instances in the image and represents the scene more accurately overall.}
    \label{fig:qual} 
\end{figure*}
\subsubsection{Augmented CLIP Bounding Box Guidance}\label{agbbox}
Based on CLIP Bounding Box guidance, we propose an Augmented CLIP Bounding Box Guidance by strengthening the guidance gradient with gradient of Gaussian noise. The aim of Augmented CLIP Bounding Box Guidance is to increase the guidance in the region where the object should appear. Along with calculating the gradient of a Gaussian Noise padded image above, we also calculate the gradient of a pure Gaussian Noise $\gamma $ with the same size. 
\begin{equation}
    \nabla L_{gauss} = -\frac{\partial L_{gauss}}{\partial \gamma}
\end{equation}
The formulation for this Augmented CLIP Bounding Box Guidance $\nabla L_{aug,boxg} $ is:
\begin{equation}  
\nabla L_{aug,boxg}  = \lambda * (\nabla L_{boxg} - \nabla L_{gaus})+ \nabla L_{gaus}     
\end{equation} 
where $\nabla L_{bbox}$ represents the gradient of Gaussian Noise padded bounding box, $\nabla  L_{gaus}$ represents the gradient of pure Gaussian Noise with the same size, $\lambda$ is a hyperparameter to control the intensity of guidance. If $\lambda$ is set to 1, Augmented CLIP Bounding Box Guidance is exactly the same as the vanilla CLIP Bounding Box Guidance. 

\subsubsection{Segmentation Map Guidance} \label{diffinit}
Given the first-stage autoencoder of the LDM \cite{rombach2022high} , $T(.)$ , segmentation map $s$ for the image and generated image $x$ in the reverse process of the diffusion model, we calculate the score to measure how  semantically close the segmentation map and the generated image are. The score is formulated as:
\begin{equation}  
    L_{segg} = T(s)*T(x)     
\end{equation} ,  the gradient of the score function for LDM \cite{rombach2022high} with respect to the latent space $z$ 
will be 
\begin{equation}  
    \nabla L_{segg} = -\frac{\partial L_{segg}}{\partial z}
\end{equation}

Then, the total diffusion guidance gradient is computed as follows:
\begin{equation}
    \nabla L_{diff} = \nabla L_{segg} + \nabla L_{text} + \nabla L_{aug,boxg}
\end{equation}

\begin{table*}[!ht]
    \caption{\textbf{Comparison against SOTA on COCO stuff \cite{caesar2018coco}}. We present the results on $64 \times 64$, and $256 \times 256$ resolutions. We present the results of different methods with different generator architectures. The models identified by \textbf{Pred} use predicted bounding box, segmentation map, or scene layouts, while \textbf{GT} identifies experiments with ground truth information.}
    \centering
    \resizebox{\textwidth}{!}{
    \begin{tabular}{l|c|c|ccccc}
    \hline
    Method & Pred / GT & Resolution & IS $\uparrow$  & FID $\downarrow$  & KID ($\times 10^2$) $\downarrow$ & SOA-C ($\times 10^2$) $\uparrow$ & SOA-I  ($\times 10^2$) $\uparrow$ \\ \hline 
    SG2Im \cite{johnson2018image} & GT & $64 \times 64$ & 5.30 & 113.61 & 58\xpm{0.3} & - & -  \\ \hline 
    \methodName{} (Ours) & GT & $64 \times 64$ & \textbf{9.05} & \textbf{67.51} & \textbf{7.86\xpm{0.087}}  & - & - \\ \hline \hline 
    SG2Im \cite{johnson2018image}& GT & $256 \times 256$ & 6.6 & 127.0 & -  & - & -\\ 
    PasteGAN \cite{li2019pastegan} &GT & $256 \times 256$ & 11.0 & 70.2 & - & - & -\\
    Specifying \cite{ashual2019specifying}  &GT & $256 \times 256$ & 12.4 & 65.2 & -  & - & -\\ 
    Canonical \cite{herzig2020learning} &GT & $256 \times 256$ & 19.5 &  64.65  & 7.03\xpm{0.177}   & 33.94 & 48.55\\ 
    LDM \cite{rombach2022high} &GT & $256 \times 256$ & \textbf{22.24} & 63.83 & 6.06\xpm{0.114}  & 45.38 & 57.22 \\ 
    \hline
    \methodName{} (Ours) & GT & $256 \times 256$ & 21.72 &63.05 & 5.54\xpm{0.105}   & 45.67	& 56.91 \\ 
    \methodName{} (Ours + Seg) & GT & $256 \times 256$ & 21.50 & \textbf{62.38} & \textbf{5.10}\xpm{0.095}  & \textbf{45.80}	& \textbf{57.39}\\  
     \hline  \hline
    Canonical \cite{herzig2020learning} &Pred & $256 \times 256$ & 9.03 &  113.30 & 7.67\xpm{0.173}  & 34.78 & 50.93\\  \hline 
    \methodName{} (Ours) & Pred & $256 \times 256$ & \textbf{22.16} & \textbf{63.27} & \textbf{4.98}\xpm{0.101}   & \textbf{43.80} & \textbf{56.61}\\ 

    \hline 
    \end{tabular} 
    }
    \label{tab:comparison_coco}
\end{table*}


\begin{table}[t]
    \caption{\textbf{Comparison against SOTA on Visual Genome \cite{krishna2017visual}.} The results are presented on images with $256 \times 256$ resolution. Our final model is the combination of predicted bounding box, with augmented bounding box guidance.}
    \centering
    \begin{tabular}{l|ccc}
    \hline
    Method & IS $\uparrow$  & FID $\downarrow$  & KID ($\times 10^2$) $\downarrow$ \\ \hline 
    Canonical \cite{herzig2020learning} & 16.5 &  45.7 & - \\ 
    LDM \cite{rombach2022high} & 20.02 & 42.69 & 8.63\xpm{0.505} \\ \hline 
    \methodName{} (Ours) & \textbf{20.25} & \textbf{42.21} & \textbf{8.43}\xpm{0.517}  \\ \hline 
    \end{tabular}
    \label{tab:comparison_vg}
\end{table}

\begin{table} 
    \centering
    \caption{\textbf{Ablation Study on COCO stuff \cite{caesar2018coco}.} We study the different components of our model. We analyze the effect of bounding box and segmentation guidance, and the different values for the bounding box guidance scale. \textbf{B}: Bounding Box, \textbf{S}: Segmentation Map.}
    \begin{tabular}{c|c|ccc} 
    \hline
    \thead{Guidance} & $\lambda$ & IS $\uparrow$ & FID $\downarrow$ & KID ($\times 10^2$) $\downarrow$  \\ \hline 
    \multicolumn{5}{c}{GT} \\ \hline
    - & - & 22.24\xpm{1.778} & 63.83 & 6.06\xpm{0.114}  \\ \hline 
    B & 1 & 21.46\xpm{1.49} & 63.14 & 5.82\xpm{0.114} \\ 
    B & 1.1 & 21.93\xpm{1.44} & 63.14 & \underline{5.21}\xpm{0.103} \\ 
    B & 1.2 & 21.72\xpm{1.45} & \underline{63.05} & 5.54\xpm{0.105} \\ 
    B & 1.3 & 21.87\xpm{1.60} & 64.19 & 5.78\xpm{0.109} \\ 
    B & 1.4  & \textbf{22.07}\xpm{1.83} & 63.76 & 5.93\xpm{0.111} \\ 
    \hline
    B + S & - & 21.50\xpm{1.31} & \textbf{62.38} & \textbf{5.10}\xpm{0.095}  \\ \hline 
    \multicolumn{5}{c}{Pred} \\ \hline
    B & - & 21.73\xpm{1.65} & 63.61 & 5.88\xpm{0.114} \\ 
    B + CLIP & - & 22.04\xpm{2.19} & 63.37 & 5.31\xpm{0.112} \\ 
    B + S + CLIP & - & \textbf{22.16}\xpm{1.65} & \textbf{63.27} & \textbf{4.98}\xpm{0.101} \\  \hline 
    \end{tabular}
    \label{tab:ablation}
\end{table}

\section{Experiments} \label{sec:res}
In this section, we present the implementation details of our method and the results of our experiments on two public benchmarks, which are commonly used for image generation from scene graphs, namely Visual Genome \cite{krishna2017visual}, and COCO stuff \cite{caesar2018coco}. We evaluate our model both quantitatively and qualitatively on these datasets and compare them against the state-of-the-art in Text2Image and Scene Graph to image models. 
\subsection{Datasets}
The Visual Genome \cite{krishna2017visual} dataset consists of images and semantic scene graph annotations, along with the bounding box coordinates. The relationships in the scene graphs of VG dataset are purely semantic and only implicitly encode geometric information; while the COCO \cite{caesar2018coco} dataset does not originally include scene graph annotations, the bounding box coordinates and the captions in this dataset are used to generate geometric scene graphs. The COCO dataset includes images with semantic segmentation, bounding box coordinates and captions as annotations.

\subsection{Experimental Setup}
For all the experiments unless specified, we use a pre-trained U-Net \cite{ronneberger2015u} as our diffusion model based on LDM \cite{rombach2022high}, and perform image generation using our proposed guidance. The diffusion model is pre-trained on the ImageNet dataset \cite{deng2009imagenet}. Our model does not require any fine-tuning and is applicable to existing networks while the guidance happens during the inference time. We adopt the LDM-8 (KL) pre-trained model. The sampling process is done via DDIM \cite{song2020denoising} sampling with 100 sampling steps. For the model trained on $64 \times 64$ images from COCO, we combine our guidance with GLIDE \cite{nichol2021glide}. 

We report the performance of our model using inception score (IS), Fr\'echet Inception Distance (FID) and Kernel Inception Distance (KID) which are common image quality metrics. In addition, we report the Semantic Object Accuracy \cite{hinz2020semantic} metrics (SOA-O and SOA-I) for our model and the LDM \cite{rombach2022high} that checks whether a pre-trained object detection model recognizes the given objects. We use CLIP \cite{radford2021learning} as the text encoder. We empirically found $0.5$ as the best value for scaling segmentation guidance in the total guidance. The architecture details will be provided in the supplementary material.

Since the VG dataset does not include semantic segmentation annotations, we omit the scene graph to segmentation step in our model and only predict the bounding box coordinates. For the same reason, our final model in VG only uses CLIP embeddings and bounding box guidance.
\subsection{Results}
We provide qualitative and quantitative results of our approach compared against the state-of-the-art. We present two variations of our model in \autoref{tab:comparison_coco} and \autoref{tab:comparison_vg}, which are either with predicted or ground truth bounding box and segmentation map information. We present more qualitative results on COCO and VG in the supplementary material.
\paragraph{Comparison against SOTA}
The results of our model compared against the state-of-the-art on COCO \cite{caesar2018coco} and Visual Genome (VG) \cite{krishna2017visual} datasets are provided in \autoref{tab:comparison_coco} and \autoref{tab:comparison_vg} respectively. As it can be seen, our proposed model \methodName{}, outperforms the state-of-the-art diffusion model, LDM \cite{rombach2022high} as well as the scene graph to image \cite{johnson2018image} model on both datasets. 

We also present some qualitative results on COCO in \autoref{fig:qual}. The qualitative results show that our proposed model generates more accurate images conditioned on the prompt. One main advantage of our model is in situations, where the number of object instances are defined. In such cases, the text guided image generation models fail in representing the scene correctly, while \methodName{} generates a more representative image.

\paragraph{Ablation Study}
We present an ablation study of the components of our model in \autoref{tab:ablation}. We analyze different values for scaling the augmented bounding box in the diffusion process, and we find the best value of $1.2$ based on FID. The best overall performance is obtained by combining the bounding box and segmentation guidance with GT values. We also analyze the effect of incorporating CLIP embbedings in the graph nodes for the models with predicted bounding box and segmentation map and show its effectiveness in improving the image generation quality. In addition, we measure the bounding box prediction error with and without using CLIP embeddings for the nodes in the graph. The bbox prediction error is $0.736$, and $0.749$, with an without CLIP embeddings respectively.

\subsection{Discussion}
The introduction of bounding box and segmentation map guidance in our approach enables the model to accurately represent the scene. As it can be seen in the qualitative results in \autoref{fig:qual}, the generated images by our model represent the input prompt more accurately. Specifically, when the input prompt defines the number of objects in the scene, previous works fail to correctly generate the specified number of objects (e.g. Generating an image of two skiers instead of one or an image of one animal instead of two). Our model can be used either with predicted bounding box and segmentation map information from a text prompt or with ground truth bounding box and segmentation map. Both variations outperform the state-of-the-art in text to image and scene graph to image generation.

\subsection{Limitations}
Despite the high capacity of our method in generation of accurate and high quality images, it still fails to generate high quality images of complex structures such as faces. This limitation is consistent in different models and has also been existing in previous work. We believe that, by fine-tuning the model on a more constrained dataset of, for example, faces, this issue can be solved. Another issue is the high time consumption for the generation of image in the reverse sampling process, which is common in diffusion models. One limitation of our method is that the guidance process requires predicted segmentation maps and bounding box information, which can be tackled by employing off-the-shelf semantic segmentation and object detection models. 
\section{Conclusion} \label{sec:conclusion}
In this work, we presented a novel guidance for the sampling process in a diffusion model. Our proposed guidance enforces geometric constraints in the sampling process using the bounding box and segmentation information predicted from a scene graph. To improve the prediction of bounding box and segmentation map from the scene graph, we encode the nodes with CLIP embedding. Our proposed guidance, as well as the employment of CLIP embeddings in the graph nodes, facilitate the generation of higher quality and more accurate images. The proposed guidance does not require any training and is applicable during the sampling process. Our method achieves better performance compared to the models trained for conditional scene graph to image generation without any training on the target datasets and also outperforms the state-of-the-art in text to image generation.

{\small
\bibliographystyle{ieee_fullname}
\bibliography{egbib}

\begin{thebibliography}{10}\itemsep=-1pt

\bibitem{alon2020bottleneck}
Uri Alon and Eran Yahav.
\newblock On the bottleneck of graph neural networks and its practical
  implications.
\newblock {\em arXiv preprint arXiv:2006.05205}, 2020.

\bibitem{ashual2019specifying}
Oron Ashual and Lior Wolf.
\newblock Specifying object attributes and relations in interactive scene
  generation.
\newblock In {\em ICCV}, 2019.

\bibitem{caesar2018coco}
Holger Caesar, Jasper Uijlings, and Vittorio Ferrari.
\newblock Coco-stuff: Thing and stuff classes in context.
\newblock In {\em Proceedings of the IEEE conference on computer vision and
  pattern recognition}, pages 1209--1218, 2018.

\bibitem{chen2017photographic}
Qifeng Chen and Vladlen Koltun.
\newblock Photographic image synthesis with cascaded refinement networks.
\newblock In {\em ICCV}, pages 1511--1520, 2017.

\bibitem{deng2009imagenet}
Jia Deng, Wei Dong, Richard Socher, Li-Jia Li, Kai Li, and Li Fei-Fei.
\newblock Imagenet: A large-scale hierarchical image database.
\newblock In {\em 2009 IEEE conference on computer vision and pattern
  recognition}, pages 248--255. Ieee, 2009.

\bibitem{dhamo2020semantic}
Helisa Dhamo, Azade Farshad, Iro Laina, Nassir Navab, Gregory~D Hager, Federico
  Tombari, and Christian Rupprecht.
\newblock Semantic image manipulation using scene graphs.
\newblock In {\em Proceedings of the IEEE/CVF Conference on Computer Vision and
  Pattern Recognition}, pages 5213--5222, 2020.

\bibitem{dhariwal2021diffusion}
Prafulla Dhariwal and Alexander Nichol.
\newblock Diffusion models beat gans on image synthesis.
\newblock {\em Advances in Neural Information Processing Systems},
  34:8780--8794, 2021.

\bibitem{esser2020taming}
Patrick Esser, Robin Rombach, and Björn Ommer.
\newblock Taming transformers for high-resolution image synthesis, 2020.

\bibitem{gafni2022make}
Oran Gafni, Adam Polyak, Oron Ashual, Shelly Sheynin, Devi Parikh, and Yaniv
  Taigman.
\newblock Make-a-scene: Scene-based text-to-image generation with human priors.
\newblock In {\em Computer Vision--ECCV 2022: 17th European Conference, Tel
  Aviv, Israel, October 23--27, 2022, Proceedings, Part XV}, pages 89--106.
  Springer, 2022.

\bibitem{garg2021unconditional}
Sarthak Garg, Helisa Dhamo, Azade Farshad, Sabrina Musatian, Nassir Navab, and
  Federico Tombari.
\newblock Unconditional scene graph generation.
\newblock In {\em Proceedings of the IEEE/CVF International Conference on
  Computer Vision}, pages 16362--16371, 2021.

\bibitem{goodfellow2014generative}
Ian Goodfellow, Jean Pouget-Abadie, Mehdi Mirza, Bing Xu, David Warde-Farley,
  Sherjil Ozair, Aaron Courville, and Yoshua Bengio.
\newblock Generative adversarial nets.
\newblock {\em Advances in neural information processing systems}, 27, 2014.

\bibitem{han2016seq}
Wei Han, Pooya Khorrami, Tom~Le Paine, Prajit Ramachandran, Mohammad
  Babaeizadeh, Honghui Shi, Jianan Li, Shuicheng Yan, and Thomas~S Huang.
\newblock Seq-nms for video object detection.
\newblock {\em arXiv preprint arXiv:1602.08465}, 2016.

\bibitem{herzig2020learning}
Roei Herzig, Amir Bar, Huijuan Xu, Gal Chechik, Trevor Darrell, and Amir
  Globerson.
\newblock Learning canonical representations for scene graph to image
  generation.
\newblock In {\em European Conference on Computer Vision}, pages 210--227.
  Springer, 2020.

\bibitem{herzig2018mapping}
Roei Herzig, Moshiko Raboh, Gal Chechik, Jonathan Berant, and Amir Globerson.
\newblock Mapping images to scene graphs with permutation-invariant structured
  prediction.
\newblock In {\em NeurIPS}, 2018.

\bibitem{hessel2021clipscore}
Jack Hessel, Ari Holtzman, Maxwell Forbes, Ronan~Le Bras, and Yejin Choi.
\newblock Clipscore: A reference-free evaluation metric for image captioning.
\newblock {\em arXiv preprint arXiv:2104.08718}, 2021.

\bibitem{hinz2020semantic}
Tobias Hinz, Stefan Heinrich, and Stefan Wermter.
\newblock Semantic object accuracy for generative text-to-image synthesis.
\newblock {\em IEEE transactions on pattern analysis and machine intelligence},
  44(3):1552--1565, 2020.

\bibitem{ho2020denoising}
Jonathan Ho, Ajay Jain, and Pieter Abbeel.
\newblock Denoising diffusion probabilistic models.
\newblock {\em Advances in Neural Information Processing Systems},
  33:6840--6851, 2020.

\bibitem{hong2018learning}
Seunghoon Hong, Xinchen Yan, Thomas~E Huang, and Honglak Lee.
\newblock Learning hierarchical semantic image manipulation through structured
  representations.
\newblock In {\em Advances in Neural Information Processing Systems}, pages
  2713--2723, 2018.

\bibitem{hu2013patchnet}
Shi-Min Hu, Fang-Lue Zhang, Miao Wang, Ralph~R Martin, and Jue Wang.
\newblock Patchnet: A patch-based image representation for interactive
  library-driven image editing.
\newblock {\em ACM Transactions on Graphics (TOG)}, 32(6):1--12, 2013.

\bibitem{isola2017image}
Phillip Isola, Jun-Yan Zhu, Tinghui Zhou, and Alexei~A Efros.
\newblock Image-to-image translation with conditional adversarial networks.
\newblock In {\em Proceedings of the IEEE conference on computer vision and
  pattern recognition}, pages 1125--1134, 2017.

\bibitem{ivgi2021scene}
Maor Ivgi, Yaniv Benny, Avichai Ben-David, Jonathan Berant, and Lior Wolf.
\newblock Scene graph to image generation with contextualized object layout
  refinement.
\newblock In {\em 2021 IEEE International Conference on Image Processing
  (ICIP)}, pages 2428--2432. IEEE, 2021.

\bibitem{ji2020action}
Jingwei Ji, Ranjay Krishna, Li Fei-Fei, and Juan~Carlos Niebles.
\newblock Action genome: Actions as compositions of spatio-temporal scene
  graphs.
\newblock In {\em Proceedings of the IEEE/CVF Conference on Computer Vision and
  Pattern Recognition}, pages 10236--10247, 2020.

\bibitem{johnson2018image}
Justin Johnson, Agrim Gupta, and Li Fei-Fei.
\newblock Image generation from scene graphs.
\newblock In {\em CVPR}, 2018.

\bibitem{johnson15}
J. {Johnson}, R. {Krishna}, M. {Stark}, L. {Li}, D.~A. {Shamma}, M.~S.
  {Bernstein}, and L. {Fei-Fei}.
\newblock Image retrieval using scene graphs.
\newblock In {\em CVPR}, 2015.

\bibitem{karras2021alias}
Tero Karras, Miika Aittala, Samuli Laine, Erik H{\"a}rk{\"o}nen, Janne
  Hellsten, Jaakko Lehtinen, and Timo Aila.
\newblock Alias-free generative adversarial networks.
\newblock {\em arXiv preprint arXiv:2106.12423}, 2021.

\bibitem{karras2020analyzing}
Tero Karras, Samuli Laine, Miika Aittala, Janne Hellsten, Jaakko Lehtinen, and
  Timo Aila.
\newblock Analyzing and improving the image quality of stylegan.
\newblock In {\em Proceedings of the IEEE/CVF Conference on Computer Vision and
  Pattern Recognition}, pages 8110--8119, 2020.

\bibitem{krishna2017visual}
Ranjay Krishna, Yuke Zhu, Oliver Groth, Justin Johnson, Kenji Hata, Joshua
  Kravitz, Stephanie Chen, Yannis Kalantidis, Li-Jia Li, David~A Shamma, et~al.
\newblock Visual genome: Connecting language and vision using crowdsourced
  dense image annotations.
\newblock {\em IJCV}, 123(1):32--73, 2017.

\bibitem{li2022cross}
Lei Li, Kai Fan, and Chun Yuan.
\newblock Cross-modal representation learning and relation reasoning for
  bidirectional adaptive manipulation.
\newblock In Lud~De Raedt, editor, {\em Proceedings of the Thirty-First
  International Joint Conference on Artificial Intelligence, {IJCAI-22}}, pages
  3222--3228. International Joint Conferences on Artificial Intelligence
  Organization, 7 2022.
\newblock Main Track.

\bibitem{li2018deeper}
Qimai Li, Zhichao Han, and Xiao-Ming Wu.
\newblock Deeper insights into graph convolutional networks for semi-supervised
  learning.
\newblock In {\em Thirty-Second AAAI conference on artificial intelligence},
  2018.

\bibitem{li2019storygan}
Yitong Li, Zhe Gan, Yelong Shen, Jingjing Liu, Yu Cheng, Yuexin Wu, Lawrence
  Carin, David Carlson, and Jianfeng Gao.
\newblock Storygan: A sequential conditional gan for story visualization.
\newblock In {\em Proceedings of the IEEE/CVF Conference on Computer Vision and
  Pattern Recognition}, pages 6329--6338, 2019.

\bibitem{li2019pastegan}
Yikang Li, Tao Ma, Yeqi Bai, Nan Duan, Sining Wei, and Xiaogang Wang.
\newblock Pastegan: A semi-parametric method to generate image from scene
  graph.
\newblock {\em Advances in Neural Information Processing Systems}, 32, 2019.

\bibitem{li2017scene}
Yikang Li, Wanli Ouyang, Bolei Zhou, Kun Wang, and Xiaogang Wang.
\newblock Scene graph generation from objects, phrases and region captions.
\newblock In {\em ICCV}, pages 1261--1270, 2017.

\bibitem{ling2021editgan}
Huan Ling, Karsten Kreis, Daiqing Li, Seung~Wook Kim, Antonio Torralba, and
  Sanja Fidler.
\newblock Editgan: High-precision semantic image editing.
\newblock {\em arXiv preprint arXiv:2111.03186}, 2021.

\bibitem{luo2021moma}
Zelun Luo, Wanze Xie, Siddharth Kapoor, Yiyun Liang, Michael Cooper,
  Juan~Carlos Niebles, Ehsan Adeli, and Fei-Fei Li.
\newblock Moma: Multi-object multi-actor activity parsing.
\newblock {\em Advances in Neural Information Processing Systems},
  34:17939--17955, 2021.

\bibitem{meng2021sdedit}
Chenlin Meng, Yutong He, Yang Song, Jiaming Song, Jiajun Wu, Jun-Yan Zhu, and
  Stefano Ermon.
\newblock Sdedit: Guided image synthesis and editing with stochastic
  differential equations.
\newblock In {\em International Conference on Learning Representations}, 2021.

\bibitem{mirza2014conditional}
Mehdi Mirza and Simon Osindero.
\newblock Conditional generative adversarial nets.
\newblock {\em arXiv preprint arXiv:1411.1784}, 2014.

\bibitem{mittal2019interactive}
Gaurav Mittal, Shubham Agrawal, Anuva Agarwal, Sushant Mehta, and Tanya Marwah.
\newblock Interactive image generation using scene graphs.
\newblock {\em arXiv preprint arXiv:1905.03743}, 2019.

\bibitem{mokady2021clipcap}
Ron Mokady, Amir Hertz, and Amit~H Bermano.
\newblock Clipcap: Clip prefix for image captioning.
\newblock {\em arXiv preprint arXiv:2111.09734}, 2021.

\bibitem{nazeri2019edgeconnect}
Kamyar Nazeri, Eric Ng, Tony Joseph, Faisal Qureshi, and Mehran Ebrahimi.
\newblock Edgeconnect: Structure guided image inpainting using edge prediction.
\newblock In {\em The IEEE International Conference on Computer Vision (ICCV)
  Workshops}, Oct 2019.

\bibitem{newell2017pixels}
Alejandro Newell and Jia Deng.
\newblock Pixels to graphs by associative embedding.
\newblock In {\em NeurIPS}, pages 2171--2180, 2017.

\bibitem{nichol2021glide}
Alex Nichol, Prafulla Dhariwal, Aditya Ramesh, Pranav Shyam, Pamela Mishkin,
  Bob McGrew, Ilya Sutskever, and Mark Chen.
\newblock Glide: Towards photorealistic image generation and editing with
  text-guided diffusion models.
\newblock {\em arXiv preprint arXiv:2112.10741}, 2021.

\bibitem{ntavelis2020sesame}
Evangelos Ntavelis, Andr{\'{e}}s Romero, Iason Kastanis, Luc {Van Gool}, and
  Radu Timofte.
\newblock {SESAME: Semantic Editing of Scenes by Adding, Manipulating or
  Erasing Objects}.
\newblock In Andrea Vedaldi, Horst Bischof, Thomas Brox, and Jan-Michael Frahm,
  editors, {\em Computer Vision -- ECCV 2020}, pages 394--411, Cham, 2020.
  Springer International Publishing.

\bibitem{oono2019graph}
Kenta Oono and Taiji Suzuki.
\newblock Graph neural networks exponentially lose expressive power for node
  classification.
\newblock {\em arXiv preprint arXiv:1905.10947}, 2019.

\bibitem{park2019semantic}
Taesung Park, Ming-Yu Liu, Ting-Chun Wang, and Jun-Yan Zhu.
\newblock Semantic image synthesis with spatially-adaptive normalization.
\newblock In {\em Proceedings of the IEEE/CVF conference on computer vision and
  pattern recognition}, pages 2337--2346, 2019.

\bibitem{pathak2016context}
Deepak Pathak, Philipp Krahenbuhl, Jeff Donahue, Trevor Darrell, and Alexei~A
  Efros.
\newblock Context encoders: Feature learning by inpainting.
\newblock In {\em CVPR}, 2016.

\bibitem{qi2018attentive}
Mengshi Qi, Weijian Li, Zhengyuan Yang, Yunhong Wang, and Jiebo Luo.
\newblock Attentive relational networks for mapping images to scene graphs.
\newblock {\em arXiv preprint arXiv:1811.10696}, 2018.

\bibitem{radford2021learning}
Alec Radford, Jong~Wook Kim, Chris Hallacy, Aditya Ramesh, Gabriel Goh,
  Sandhini Agarwal, Girish Sastry, Amanda Askell, Pamela Mishkin, Jack Clark,
  et~al.
\newblock Learning transferable visual models from natural language
  supervision.
\newblock In {\em International Conference on Machine Learning}, pages
  8748--8763. PMLR, 2021.

\bibitem{ramesh2022hierarchical}
Aditya Ramesh, Prafulla Dhariwal, Alex Nichol, Casey Chu, and Mark Chen.
\newblock Hierarchical text-conditional image generation with clip latents.
\newblock {\em arXiv preprint arXiv:2204.06125}, 2022.

\bibitem{rombach2022high}
Robin Rombach, Andreas Blattmann, Dominik Lorenz, Patrick Esser, and Bj{\"o}rn
  Ommer.
\newblock High-resolution image synthesis with latent diffusion models.
\newblock In {\em Proceedings of the IEEE/CVF Conference on Computer Vision and
  Pattern Recognition}, pages 10684--10695, 2022.

\bibitem{ronneberger2015u}
O. Ronneberger, P.Fischer, and T. Brox.
\newblock U-net: Convolutional networks for biomedical image segmentation.
\newblock In {\em Medical Image Computing and Computer-Assisted Intervention
  (MICCAI)}, volume 9351 of {\em LNCS}, pages 234--241. Springer, 2015.
\newblock (available on arXiv:1505.04597 [cs.CV]).

\bibitem{saharia2022palette}
Chitwan Saharia, William Chan, Huiwen Chang, Chris Lee, Jonathan Ho, Tim
  Salimans, David Fleet, and Mohammad Norouzi.
\newblock Palette: Image-to-image diffusion models.
\newblock In {\em ACM SIGGRAPH 2022 Conference Proceedings}, pages 1--10, 2022.

\bibitem{sohl2015deep}
Jascha Sohl-Dickstein, Eric Weiss, Niru Maheswaranathan, and Surya Ganguli.
\newblock Deep unsupervised learning using nonequilibrium thermodynamics.
\newblock In {\em International Conference on Machine Learning}, pages
  2256--2265. PMLR, 2015.

\bibitem{song2020denoising}
Jiaming Song, Chenlin Meng, and Stefano Ermon.
\newblock Denoising diffusion implicit models.
\newblock {\em arXiv preprint arXiv:2010.02502}, 2020.

\bibitem{suhail2021energy}
Mohammed Suhail, Abhay Mittal, Behjat Siddiquie, Chris Broaddus, Jayan Eledath,
  Gerard Medioni, and Leonid Sigal.
\newblock Energy-based learning for scene graph generation.
\newblock In {\em Proceedings of the IEEE/CVF Conference on Computer Vision and
  Pattern Recognition}, pages 13936--13945, 2021.

\bibitem{sun2019image}
Wei Sun and Tianfu Wu.
\newblock Image synthesis from reconfigurable layout and style.
\newblock In {\em Proceedings of the IEEE/CVF International Conference on
  Computer Vision}, pages 10531--10540, 2019.

\bibitem{sylvain2021object}
Tristan Sylvain, Pengchuan Zhang, Yoshua Bengio, R~Devon Hjelm, and Shikhar
  Sharma.
\newblock Object-centric image generation from layouts.
\newblock In {\em Proceedings of the AAAI Conference on Artificial
  Intelligence}, volume~35, pages 2647--2655, 2021.

\bibitem{tang2020unbiased}
Kaihua Tang, Yulei Niu, Jianqiang Huang, Jiaxin Shi, and Hanwang Zhang.
\newblock Unbiased scene graph generation from biased training.
\newblock In {\em Proceedings of the IEEE/CVF Conference on Computer Vision and
  Pattern Recognition}, pages 3716--3725, 2020.

\bibitem{teng2021global}
Zhu Teng, Yani Duan, Yan Liu, Baopeng Zhang, and Jianping Fan.
\newblock Global to local: Clip-lstm-based object detection from remote sensing
  images.
\newblock {\em IEEE Transactions on Geoscience and Remote Sensing}, 60:1--13,
  2021.

\bibitem{wang2018high}
Ting-Chun Wang, Ming-Yu Liu, Jun-Yan Zhu, Andrew Tao, Jan Kautz, and Bryan
  Catanzaro.
\newblock High-resolution image synthesis and semantic manipulation with
  conditional gans.
\newblock In {\em Proceedings of the IEEE conference on computer vision and
  pattern recognition}, pages 8798--8807, 2018.

\bibitem{xu2017scenegraph}
Danfei Xu, Yuke Zhu, Christopher Choy, and Li Fei-Fei.
\newblock Scene graph generation by iterative message passing.
\newblock In {\em CVPR}, 2017.

\bibitem{xu2022hierarchical}
Xiaogang Xu and Ning Xu.
\newblock Hierarchical image generation via transformer-based sequential patch
  selection.
\newblock In {\em Proceedings of the AAAI Conference on Artificial
  Intelligence}, 2022.

\bibitem{yang2021layouttransformer}
Cheng-Fu Yang, Wan-Cyuan Fan, Fu-En Yang, and Yu-Chiang~Frank Wang.
\newblock Layouttransformer: Scene layout generation with conceptual and
  spatial diversity.
\newblock In {\em Proceedings of the IEEE/CVF conference on computer vision and
  pattern recognition}, pages 3732--3741, 2021.

\bibitem{yeh2017semantic}
Raymond~A Yeh, Chen Chen, Teck Yian~Lim, Alexander~G Schwing, Mark
  Hasegawa-Johnson, and Minh~N Do.
\newblock Semantic image inpainting with deep generative models.
\newblock In {\em CVPR}, pages 5485--5493, 2017.

\bibitem{Yu_2019_ICCV}
Jiahui Yu, Zhe Lin, Jimei Yang, Xiaohui Shen, Xin Lu, and Thomas~S. Huang.
\newblock Free-form image inpainting with gated convolution.
\newblock In {\em Proceedings of the IEEE/CVF International Conference on
  Computer Vision (ICCV)}, October 2019.

\bibitem{zellers2018neural}
Rowan Zellers, Mark Yatskar, Sam Thomson, and Yejin Choi.
\newblock Neural motifs: Scene graph parsing with global context.
\newblock In {\em CVPR}, pages 5831--5840, 2018.

\bibitem{zhang2018photographic}
Zizhao Zhang, Yuanpu Xie, and Lin Yang.
\newblock Photographic text-to-image synthesis with a hierarchically-nested
  adversarial network.
\newblock In {\em Proceedings of the IEEE conference on computer vision and
  pattern recognition}, pages 6199--6208, 2018.

\bibitem{zhao2018image}
Bo Zhao, Lili Meng, Weidong Yin, and Leonid Sigal.
\newblock Image generation from layout.
\newblock In {\em CVPR}, 2019.

\bibitem{zhu2017unpaired}
Jun-Yan Zhu, Taesung Park, Phillip Isola, and Alexei~A Efros.
\newblock Unpaired image-to-image translation using cycle-consistent
  adversarial networks.
\newblock In {\em ICCV}, 2017.

\bibitem{zia2020text}
Tehseen Zia, Shahan Arif, Shakeeb Murtaza, and Mirza~Ahsan Ullah.
\newblock Text-to-image generation with attention based recurrent neural
  networks.
\newblock {\em arXiv preprint arXiv:2001.06658}, 2020.

\end{thebibliography}
}

\end{document}